\documentclass[runningheads]{llncs}
\usepackage{graphicx}
\usepackage{array,multirow,graphicx} 
\usepackage{float}
\usepackage{booktabs}
\usepackage{subfig}
\usepackage{amsmath}
\usepackage{comment}

\begin{document}
%
\title{Forecasting Auxiliary Energy Consumption for Electric Heavy-Duty Vehicles}
%
%
%
\author{Yuantao Fan\inst{1} \and
Zhenkan Wang\inst{2} \and
Sepideh Pashami\inst{1,3} \and
S\l awomir Nowaczyk\inst{1} \and
Henrik Ydreskog\inst{2}
}
\authorrunning{Y. Fan et al.}
\institute{Center for Applied Intelligent System Research, Halmstad University \email{\{firstname.lastname\}@hh.se}
\and Volvo Group \email{\{firstname.lastname\}@volvo.com}
\and Research Institutes of Sweden \email{\{firstname.lastname\}@ri.se}
}

%
\maketitle

\begin{abstract}
Accurate energy consumption prediction is crucial for optimizing the operation of electric commercial heavy-duty vehicles, e.g., route planning for charging. Moreover, understanding why certain predictions are cast is paramount for such a predictive model to gain user trust and be deployed in practice.
Since commercial vehicles operate differently as transportation tasks, ambient, and drivers vary, 
a heterogeneous population is expected when building an AI system for forecasting energy consumption. The dependencies between the input features and the target values are expected to also differ across sub-populations.
One well-known example of such a statistical phenomenon is Simpson’s paradox. 
In this paper, we illustrate that such a setting poses a challenge for existing XAI methods that produce global feature statistics, e.g. LIME or SHAP, causing them to yield misleading results.
We demonstrate a potential solution by training multiple regression models on subsets of data. It not only leads to superior regression performance but also more relevant and consistent LIME explanations.
Given that the employed groupings correspond to relevant sub-populations, the associations between the input features and the target values are consistent within each cluster but different across clusters.
Experiments on both synthetic and real-world datasets show that such splitting of a complex problem into simpler ones yields better regression performance and interpretability.

\end{abstract}

\section{Introduction}

As the ongoing electrification of transportation solutions is accelerating, an increasing amount of new electric heavy-duty vehicles are designed, manufactured, and deployed. 
Optimizing the operation of electric commercial heavy-duty vehicles requires accurate energy consumption prediction in order to, for example, plan the route in the most efficient way and determine the best time and location for charging. Since such optimization directly impacts the operation and profitability of the vehicle fleet or logistics site, any decision made by the AI system interacts in complex ways with human planners, managers, and operators. Therefore, understanding why certain predictions are cast in any given situation is paramount for such a predictive model to gain user trust and be deployed in practice.
%

Commercial vehicles operate in many different situations and perform many different tasks. As transportation tasks, ambient conditions, and drivers vary, so do the specifics of internal processes of the driveline, as well as auxiliary components. The possible circumstances that affect all the essential subsystems, such as air conditioners, cabin heaters, energy converters, and more, are too numerous to account for explicitly.
Therefore, an AI system forecasting energy consumption needs to be able to handle a heterogeneous population. Particularly, the dependencies between the input features and the target values will vary across sub-populations. In other words, a given feature that is positively correlated with energy consumption in one setting might be uncorrelated with it, or even negatively correlated, in another setting.

In this paper, we illustrate that such a situation poses a challenge for existing XAI methods that produce global feature statistics, such as LIME or SHAP, causing them to yield misleading results.
One well-known example of such a statistical phenomenon is Simpson’s paradox, i.e., the association between two variables may emerge, reverse, or disappear when the entire dataset is divided into sub-populations.
For example, two variables may be positively correlated in the entire population but can be independent or negatively correlated in some of the sub-populations. In this case, relying only on a global model for interpreting the prediction may lead to neglecting crucial details; complete understanding can only be obtained with a local model, i.e., one trained using peers within a similar neighborhood.

In our study, we propose and evaluate a potential solution involving the training of multiple regression models on various subsets of data. 
This approach, Fleet-based Regression (FBR), has two key advantages. First, it allows us to identify relevant sub-patterns within the global regression task, leading to higher overall regression performance. Second, it uncovers more relevant patterns within the data, leading to more consistent LIME (Local Interpretable Model-agnostic Explanations) explanations.
Our method capitalizes on the idea of identifying relevant sub-populations within the overall dataset. By dividing the dataset into these groups, we ensure that the associations between the input features and the target values remain consistent within each individual group, even if they differ across separate groups. Our approach is fundamentally focused on discovering pertinent sub-patterns within the overarching regression task, which breaks down a complex problem into smaller, more manageable parts, making it easier for LIME to find suitable explanations.

We first test this approach through a preliminary experiment conducted across a number of synthetic datasets. These datasets are based on Simpson's paradox and exhibit varying levels of complexity. The results demonstrate that by partitioning a complicated issue into simpler sub-problems, we not only improved regression performance but also enhanced the interpretability of the models.
Specifically, we found that the LIME results derived from models trained on these sub-populations showcase more consistent explanations than those obtained from models trained on the entire dataset. This finding implies that our approach of segregating the data into relevant sub-populations has the potential to enhance both the performance and the interpretability of regression models. 


Furthermore, we have applied FBR to predicting auxiliary energy consumption on a dataset collected from a heterogeneous population of electric trucks, deployed and undertaking transportation tasks under different conditions, e.g. locations, ambient temperature, usages, etc. 
The dataset is comprised of aggregated features of on-board sensor data, including energy consumption of auxiliary systems.
In addition, domain experts split these vehicles into four fleets, i.e., sub-populations, based on the usage and operating conditions. 
Experimental results on forecasting the energy consumption for the heaters show that FBR with regression models built locally on each sub-population (determined via the k-means clustering algorithm), outperforms the traditional approach, yielding lower prediction error, and more consistent explanations.

\section{Related Work}

The concept of finding relevant training samples instead of using all the available data to build accurate Machine Learning models is not new in the field.
For example, k-nearest Neighbor methods take advantage of proximity to perform classification or regression based on the sub-population or peer groups the sample belongs to, given a predetermined value of neighborhood size. A different but related idea is used by tree-based regression methods that partition the data into sub-spaces and perform the prediction tasks.
This divide-and-conquer idea was applied to many application domains, e.g. regression modeling on clusters of homogeneous sub-population \cite{verbeke2019fleet,zhang2019quantile,hardegree2006predicting,li2016cluster}, or generating robust features for regression \cite{fan2020transfer} in heterogeneous populations.
In recent years, transfer learning has gained popularity among the topics in Machine Learning. Instance-based (or sample-based) transfer learning \cite{pan2010survey,Wang2018Instance} aims at selecting or reweighting relevant training samples in the source domain for use in the target domain, e.g., selecting useful samples, excluding samples that reduce model performances, or minimizing the differences in the data distribution between the source and the target domain, etc.
As an example, TrAdaBoost \cite{dai2007Boosting} proposed by Dai et al. iteratively reweights samples in the training dataset to improve the prediction performance; He et al. extended that work for multiclass classification \cite{HE2020118}.
The majority of the methods in this field focus on improving performance via reweighting samples in the source domain or refining the training set.

However, for many AI applications, especially in industrial settings, it is important to acquire explanations to interpret the reasoning behind the predictions cast by ML models.
In datasets with heterogeneous sub-populations, it is interesting to come up with human-interpretable sub-populations, or clustering results, of the dataset if applicable. 
It is also useful to provide interpretable clustering results to the field expert and receive supervision in tuning the clustering settings as a means to improve the clustering quality and, as a result, the performance in prediction.
Along those lines, Gupta et al. \cite{gupta2021deep} proposed interpretable deep clustering, selecting features on the cluster level via a gate matrix, and demonstrated the interpretability of the method on both sample and cluster levels (see~\cite{InteractiveClusteringSurvey} for an overview of the field).
Bertsimas et al. \cite{bertsimas2018interpretable,bertsimas2021interpretable} proposed the use of mixed integer optimization to generate interpretable tree-based clustering methods, leading to high clustering quality and interpretability.

\begin{figure}[t]
\begin{minipage}[h]{0.49\linewidth}\centering\includegraphics[width=1\linewidth]{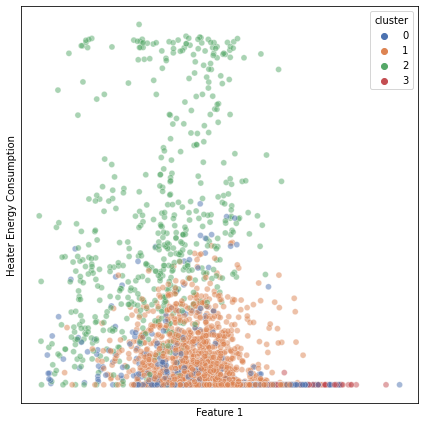}
\end{minipage}
\begin{minipage}[h]{0.49\linewidth}\centering\includegraphics[width=1\linewidth]{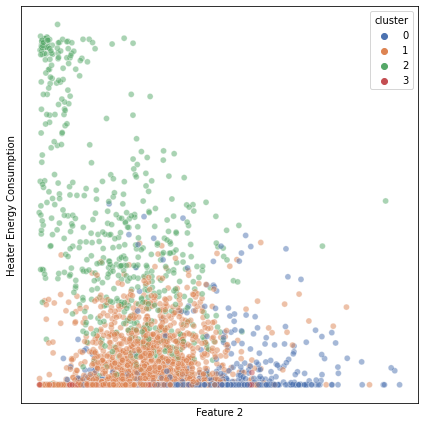}
\end{minipage}
\caption{The relation between the auxiliary energy consumption of heavy-duty vehicles and the ambient temperature}
\label{fig:p4}
\end{figure}

\section{Problem Statement}

As mentioned above, commercial vehicles operate under varying conditions and perform different tasks.
Thus, a heterogeneous population is expected when building an AI system for forecasting energy consumption.
As an example, we illustrate such heterogeneity with a real-world dataset collected from operations of commercial heavy-duty vehicles in Figure~\ref{fig:p4}, where each point represents a trip and therein aggregates the sensor data recorded, including auxiliary energy consumed.
Figure~\ref{fig:p4} shows trips of energy consumed by the heaters 
versus two aggregated features (anonymized due to secrecy requirements), with colors indicating fleets of different types, labeled by domain experts. It can be observed that the spread of cluster $2$ is wider compared with the rest of the clusters, i.e., there is heterogeneity in the dataset.
If a regression model were to be trained on the entire population versus only the sub-population of cluster $2$, the fitted model parameters, e.g., the slope and the intercept for linear models, would be very different.

To formally define the problem, let us denote a such multivariate (with $K$ features) dataset as $\mathbf{X}$, which contains $M$ sub-populations $\mathbf{X}^c$, $c = {1, 2, ..., M}$. 
Each instance, or a trip, is denoted as $\mathbf{x}_i^c \in \mathrm{R}^{K}$, where $i$ corresponds to the trip id, and $c$ indicates the sub-population $\mathbf{X}^c$ it belongs to. 
The traditional approach to predicting the target (e.g., energy consumption, denoted as $y_i$) is to train a global regression model $f$: $\mathbf{X} \rightarrow Y$, with all labeled samples $\{\mathbf{x}_i, y_i\}$ in the dataset $\mathbf{X}$.
However, when multiple distinctively different sub-populations exist in the dataset, where marginal distributions $P(X^c)$ or conditional distribution $P(Y|X^c)$ are different, using the traditional approach may result in inferior prediction accuracy and inconsistent explanations. This paper focuses on the XAI methods using feature importance, specifically LIME, but we believe the results should hold for other types of explanations as well.

FBR, performing regression on sub-populations, is essentially based on the divide-and-conquer idea. It first performs clustering, e.g., using k-means, on the entire population, assigning a cluster membership $c$ to each sample $x_i^c$. Thus, samples given the same $\hat{c}$ are considered peers in cluster $c$.
In the second step, for each cluster $\hat{c}$, a corresponding (local) regression model $f_c$: $\mathbf{x} \rightarrow y$ is trained.
In the testing phase, the test samples are first assigned to a specific cluster $\hat{c}$, and then the corresponding local regression model $f_{\hat{c}}$ is used for forecasting the target, i.e., $\hat{y_i} = f_{\hat{c}}(x_i^{\hat{c}})$.
In addition, feature statistics calculated using suitable XAI methods, such as LIME, for interpreting the prediction result will be conducted via the selected local model $f_{\hat{c}}$, without contaminating it by the outputs of other, irrelevant models.

\begin{figure}[t]
\begin{minipage}[h]{0.49\linewidth}\centering\includegraphics[width=1\linewidth]{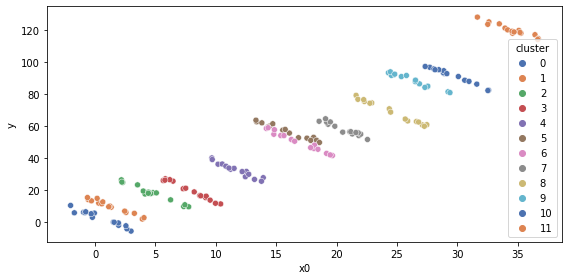} 
\end{minipage}
\hfill
\begin{minipage}[h]{0.49\linewidth}\centering\includegraphics[width=1\linewidth]{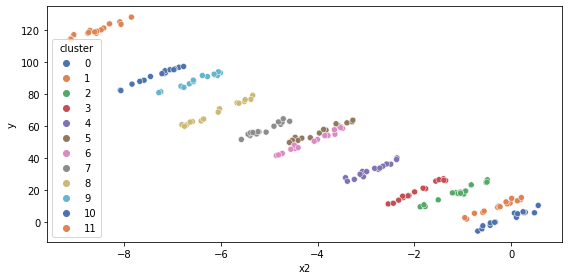}
\end{minipage}
\vfill
\begin{minipage}[h]{0.49\linewidth}\centering\includegraphics[width=1\linewidth]{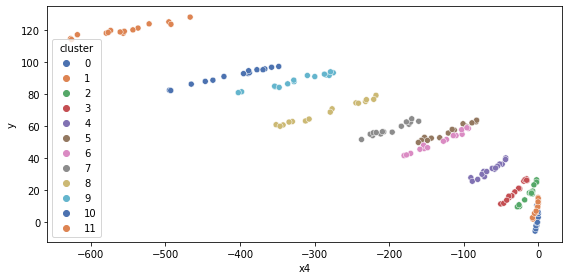}
\end{minipage}
\hfill
\vspace{-2mm}
\begin{minipage}[h]{0.49\linewidth}\centering\includegraphics[width=1\linewidth]{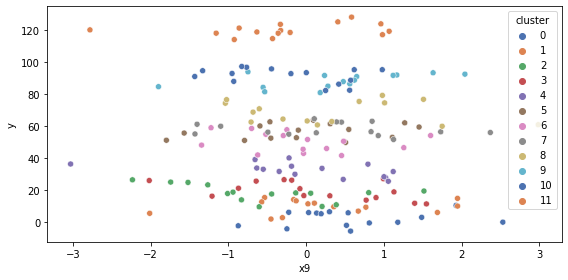}
\end{minipage}
\caption{Synthetic data capturing the Simpson's paradox idea; the four sub-plots show four (out of 10 total) input features, where each input feature has a different relation to the target variable and different cluster separability.}
\label{fig:p0}
\end{figure}

\section{Experiment}

\subsection{Synthetic Datasets}

We first demonstrate the proposed approach with a synthetic multi-dimensional dataset exhibiting the Simpson's paradox, comparing the prediction accuracy as well as the explanations stability gained versus the traditional global approach.
An illustration of the synthetic dataset used in the experimental evaluation is shown in Figure~\ref{fig:p0}. As can be seen, the association between the two variables in the entire population is different from its sub-populations (shown with different colors).
To generate such a synthetic dataset, a global linear model $f_g$ was first created, with its slope $k_g$ and intercept $b_g$ drawn from two normal distributions $\mathcal{N}_{k_g}$, and $\mathcal{N}_{b_g}$. 
Given $M$ sub-populations, center points $(y^c, x^c)$ for each sub-population are generated using $y = k_gx + b_g + \epsilon$, where $x$ is drawn from a uniform distribution over a pre-determined interval, and $\epsilon$ is random noise.

Afterward, samples in each sub-population were generated using local linear models $f_l^c$, where $c$ corresponds to a specific sub-population around the cluster center $(y_c, x_c)$. The slope $k_l^c$ and the bias $b_l^c$ for each sub-population are drawn from $\mathcal{N}_{k_l}$, and $\mathcal{N}_{b_l}$, different from the ones for the global model $\mathcal{N}_{k_g}$, and $\mathcal{N}_{b_g}$.
Each sample $(y^c_i, x^c_i)$ in cluster $c$ was then generated around the appropriate cluster center using a set of linear models $y - y_c= k_l^c(x-x_c) + b_l^c + \epsilon$.
Due to the differences between $\mathcal{N}_{k_l}$ and $\mathcal{N}_{g_l}$, real-valued pairs $(x, y)$ capturing the Simpson's paradox is generated, as shown in the top-left subplot of Figure~\ref{fig:p0}.
In order to extend the dataset into higher dimensions, a set of linearly and quadratically (shown in Figure~\ref{fig:p0} top-right and bottom-left subplots, respectively) correlated features to the original feature were generated. In addition, a set of random features (illustrated in Figure~\ref{fig:p0} bottom-right subplot), with no correlation to the original feature, were generated using normal distributions.

To summarize, experiments were conducted using the synthetic dataset 
containing $10$ features, including i) $x_0$, as the original feature for generating the targets; ii) ${x_1, x_2, x_3}$, as three features linearly dependent on $x_0$; iii) ${x_4, x_5, x_6}$, as three features quadratically dependent on $x_0$; iv) and finally ${x_7, x_8, x_9}$, as three features randomly drawn from a normal distribution and with no correlation to other features nor the target. The dataset was specifically designed to benefit from our divide-and-conquer approach; however, it is worth noting that even powerful classifiers such as Random Forest fail to achieve satisfactory results when applied directly.

We compare the performance of the traditional approach versus the proposed FBR in forecasting the target using all $10$ features.
For both experiments, regression models employed and evaluated include (i) Random Forest regressor with $100$ estimators; (ii) Ridge regression with L2 regularization and an alpha of $1$; (iiii) k-NearestNeighbors (kNN) regressor with a $k$ equal to $5$. 
The performance was measured with four metrics, i.e., mean absolute error (MAE), mean squared error (MSE), r-squared coefficient, and mean absolute percentage error (MAPE).
The experiments were conducted using $4$-fold cross-validation, and for the regression models, scikit-learn library \cite{scikit-learn} was employed.

The results shown in Table~\ref{tab:syn-reg} indicate that the proposed FBR approach (for all regression models) outperforms, often dramatically, the traditional approach. For the traditional approach, RF and kNN perform slightly better than the rest of the regression models since RF is a tree-based method that partitions data into sub-spaces and learns the mean of targets. At the same time, kNN seeks peers that are likely within the same sub-population, given an appropriate number of neighbors.

\begin{figure}[h]
\begin{minipage}[h]{0.49\linewidth}\centering\includegraphics[width=1\linewidth]{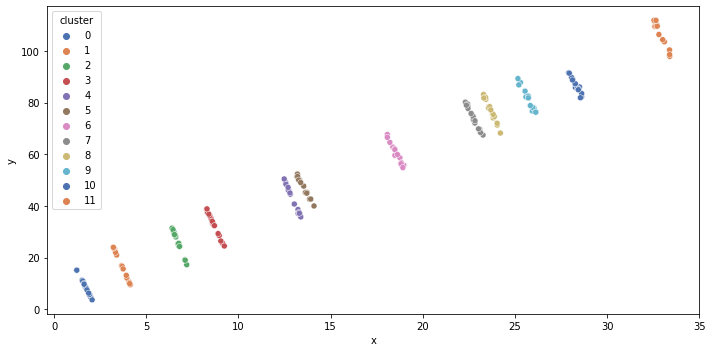} 
\end{minipage}
\hfill
\begin{minipage}[h]{0.49\linewidth}\centering\includegraphics[width=1\linewidth]{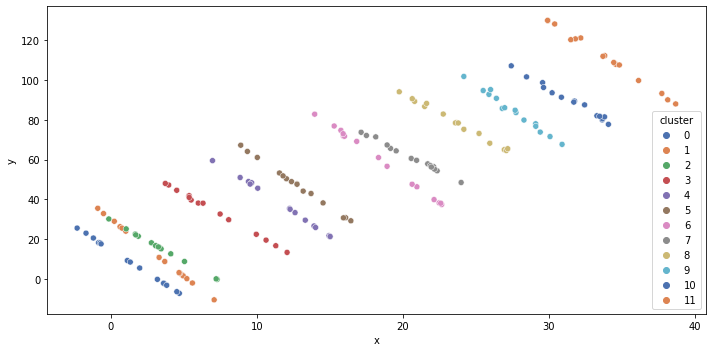} 
\end{minipage}
\caption{Synthetic data generated with different meta-parameters; the dataset on the right is designed to be more difficult for regression.}
\label{fig:p3}
\end{figure}

It is also of interest to investigate how the performance deteriorates when the dataset is generated to become more difficult for the regression task.
With different configurations (choices of meta-parameters), the synthetic dataset generated can vary over, e.g., the spread of sample points in each cluster. Intuitively, the more extensive the spread of each cluster is, the more difficult the regression task. As an example, shown in Figure~\ref{fig:p3}, samples in the left subplot have a narrower spread and a sharper slope in each sub-population, compared to the ones in the right figure.

\begin{figure}[h]
\centering
\begin{minipage}[h]{0.49\linewidth}\centering\includegraphics[width=1\linewidth]{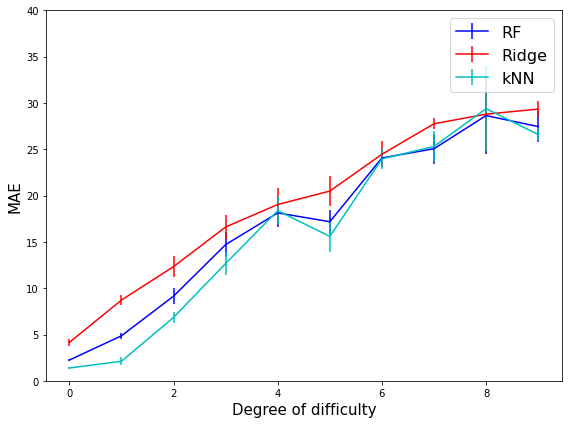}
\end{minipage}
\hfill
\begin{minipage}[h]{0.49\linewidth}\centering\includegraphics[width=1\linewidth]{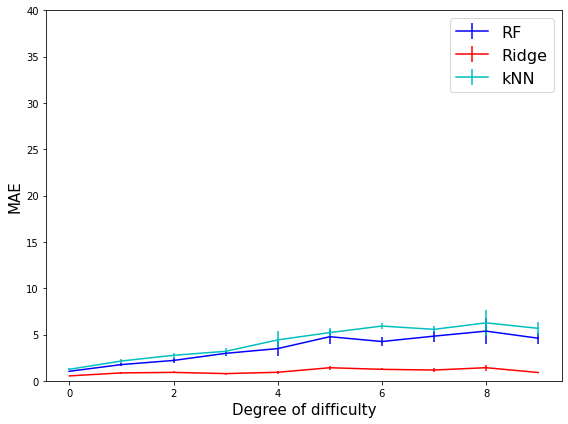}
\end{minipage}
\caption{Regression performance comparison between the traditional approach (left) and FBR (right), as the task difficulty increases.}
\label{fig:RegComparison}
\end{figure}

The corresponding results are presented in Figure~\ref{fig:RegComparison}, where the left subplot illustrates the performance deterioration using the traditional approach. On the other hand, the proposed approach deteriorates significantly less, as seen in the right-hand side subplot. Since the data were initially generated using a linear model, with an apparent clustering of the sub-population, the ridge regression achieved the best performance. RF and kNN follow it. 

\begin{table*}[t]
\hspace{-1cm}
\caption{Regression performance using the FBR approach, i.e., training a separate regression model on each sub-population.}
\vspace{-0.35cm}
\begin{center}
\begin{tabular}{c|ccccc}
\toprule
 & MAE & MSE & $R^2$ & MAPE \\ \midrule
RF &  7.8321  ±  0.6006 &  113.8849  ±  14.9918 &  0.9086  ±  0.0133 &  0.3346  ±  0.1379 \\
Ridge &  9.5153  ±  0.7515 &  132.7168  ±  12.2148 &  0.8939  ±  0.007 &  0.3415  ±  0.1083 \\
kNN &  7.9276  ±  0.1627 &  115.9756  ±  2.7621 &  0.907  ±  0.0045 &  0.3363  ±  0.1759 \\
\midrule
FBR-RF &  1.511  ±  0.2285 &  3.9855  ±  1.1441 &  0.9954  ±  0.0011 &  0.094  ±  0.0745 \\
FBR-Ridge &  1.1132  ±  0.0976 &  2.0389  ±  0.163 &  0.9976  ±  0.0002 &  0.0649  ±  0.0453 \\
FBR-kNN &  1.5241  ±  0.2504 &  4.0816  ±  1.2844 &  0.9953  ±  0.0013 &  0.1008  ±  0.0783 \\
\bottomrule
\end{tabular}
\end{center}
\label{tab:syn-reg}
\end{table*}

\begin{figure}[h]
\begin{minipage}[h]{0.5\linewidth}\centering\includegraphics[width=1\linewidth]{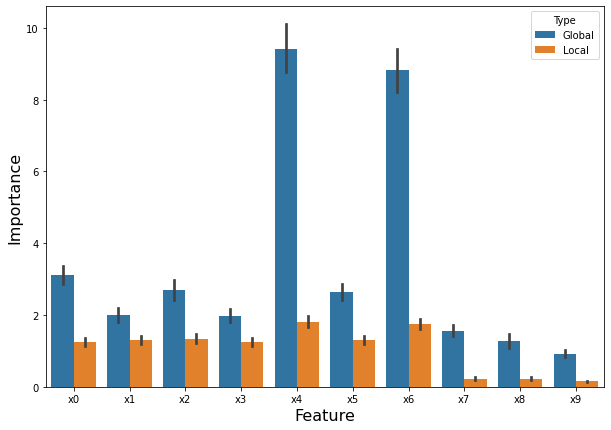}
\end{minipage}
\hfill
\begin{minipage}[h]{0.5\linewidth}\centering\includegraphics[width=1\linewidth]{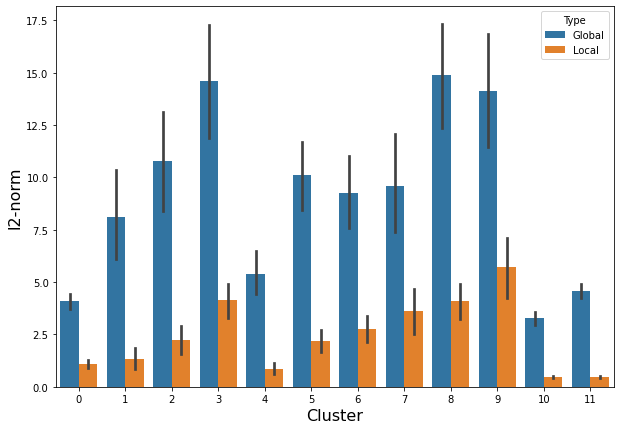}
\end{minipage}
\caption{Comparison of feature importance present in the LIME explanation (left) between the traditional approach vs. FBR; Consistency analysis in LIME explanations via inter-cluster distance measure (right)}
\label{fig:syn-exp-LIME}
\end{figure}

In order to evaluate explainability benefits, we have generated, using LIME, feature importance statistics for all $10$ features comparing both approaches, shown in Figure~\ref{fig:syn-exp-LIME} (left). 
Random forest was adopted for the regression tasks. 
For the traditional approach, one global RF model was built on the entire training population and applied to the testing samples; for FBR, RF models were trained on each cluster of samples, and testing samples were matched to the peer cluster and its corresponding local RF regression model.
As expected, features $x_7, x_8, x_9$ are of the lowest importance, as they do not correlate with the target. It is noted that the LIME feature importance of the local RF models was a lot lower compared with other features. However, this difference is less obvious for the global models.
Furthermore, features $x_0$ to $x_6$ all exhibit a fairly straightforward relation (either linear or quadratic) to the target. Intuitively, the differences between the importance of these features shall be subtle. 
It is observed that the differences between the importance of these features explained via the local model are indeed relatively small (with the only exceptions being the underappreciated features $x_4$ and $x_6$).
This weakness is greatly exacerbated by the global model, and additionally, there are two clear peaks obtained using the global model ($x_1$ and $x_3$) -- a phenomenon that is difficult to justify and indicates a serious potential to mislead human experts interpreting these results.

\begin{figure}[h]
\begin{minipage}[h]{0.5\linewidth}\centering\includegraphics[width=1\linewidth]{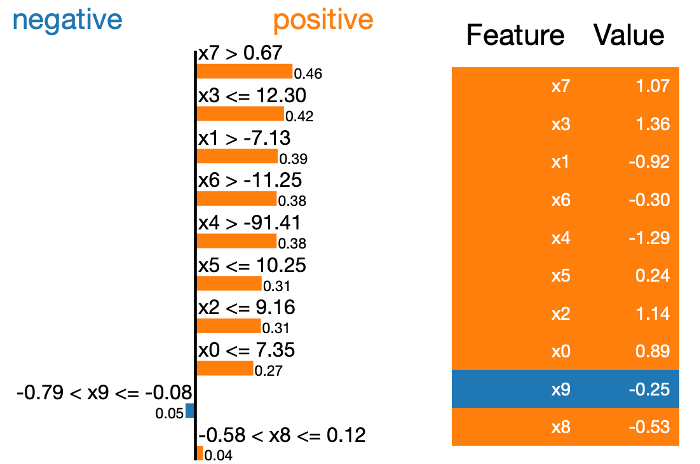}
\end{minipage}
\hfill
\begin{minipage}[h]{0.5\linewidth}\centering\includegraphics[width=1\linewidth]{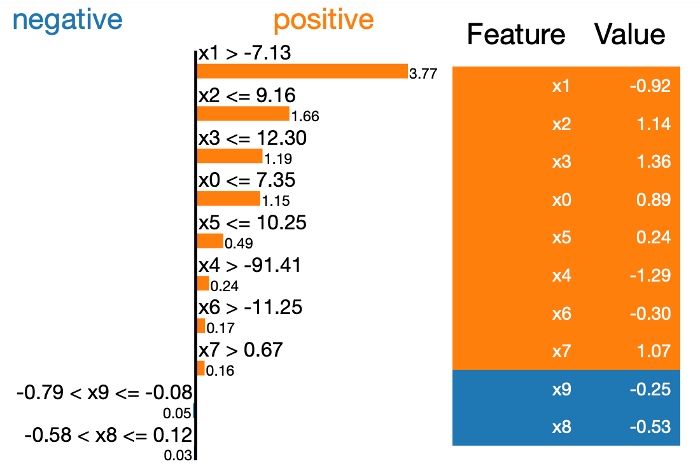}
\end{minipage}
\caption{LIME explanation on the prediction cast by a global model (left) and a local model (right)}
\label{fig:syn-exp-LIME-comp}
\end{figure}

Figure~\ref{fig:syn-exp-LIME-comp} shows LIME explanation on the prediction (of the same test sample), cast by a global model (left subplot) versus by a local model (right subplot). In this example, as is shown in the left subplot, $x_7$ was considered the most influential factor, when forecasting with a global model, which is misleading. In contrast, all three irrelevant features $x_7, x_8, x_9$ were ranked as the lowest factors in the explanation, when forecasting using the local model.

The right subplot of Figure~\ref{fig:syn-exp-LIME} shows the comparison of explanation consistency within each sub-population/cluster: the mean and the standard deviation of all pairwise distances (using $l_2$-norm) between the explanations (feature importance vectors) of samples within the same cluster.
%
%
A higher mean of all pair-wise distances in a cluster corresponds to less consistent explanations, i.e., a situation where two similar data instances (peers) are explained in more diverse ways. As shown in the figure, the explanations provided using the local regression model are significantly more consistent than the global one.

\subsection{Energy Consumption Dataset}

The energy consumption dataset was collected from $62$ electric trucks operating in several countries during a period of several months, including aggregated features over trips, computed using onboard sensor data, of vehicle mileage, speed, ambient temperature, and energy consumed for auxiliary systems.
Regression models were trained on these aggregated features to estimate the amount of energy consumed by the heaters.
Three methods were compared: i) FBR based on k-means clustering; ii) FBR based on domain expert groupings; and iii) the traditional approach, i.e., regression with a global model on the entire population.

For FBR with k-means clustering, aggregated features were standardized and utilized for both clustering and regression tasks, and the number of clusters was set to five. For each instance, local explanations were generated using LIME.
FBR using groupings suggested by the expert, employed a division into sub-populations based on the assumptions related to domain knowledge in terms of expected usage of heaters under different circumstances.
%
The regression models evaluated in our experiments include (i) Random Forest regressor with $100$ estimators; (ii) Ridge regression with L2 regularization and an alpha of $1$; (iii) k-NearestNeighbors (kNN) regressor with a $k$ equal to $5$; iv) multi-layer perception (MLP) regressor with two hidden layers, each containing $10$ ReLU units, optimized using ADAM.
The experiments were conducted using $4$-fold cross-validation, and for the regression models, scikit-learn library \cite{scikit-learn} was employed.

\begin{figure}[h]
\begin{center}
\begin{minipage}[h]{0.49\linewidth}\centering\includegraphics[width=1\linewidth]{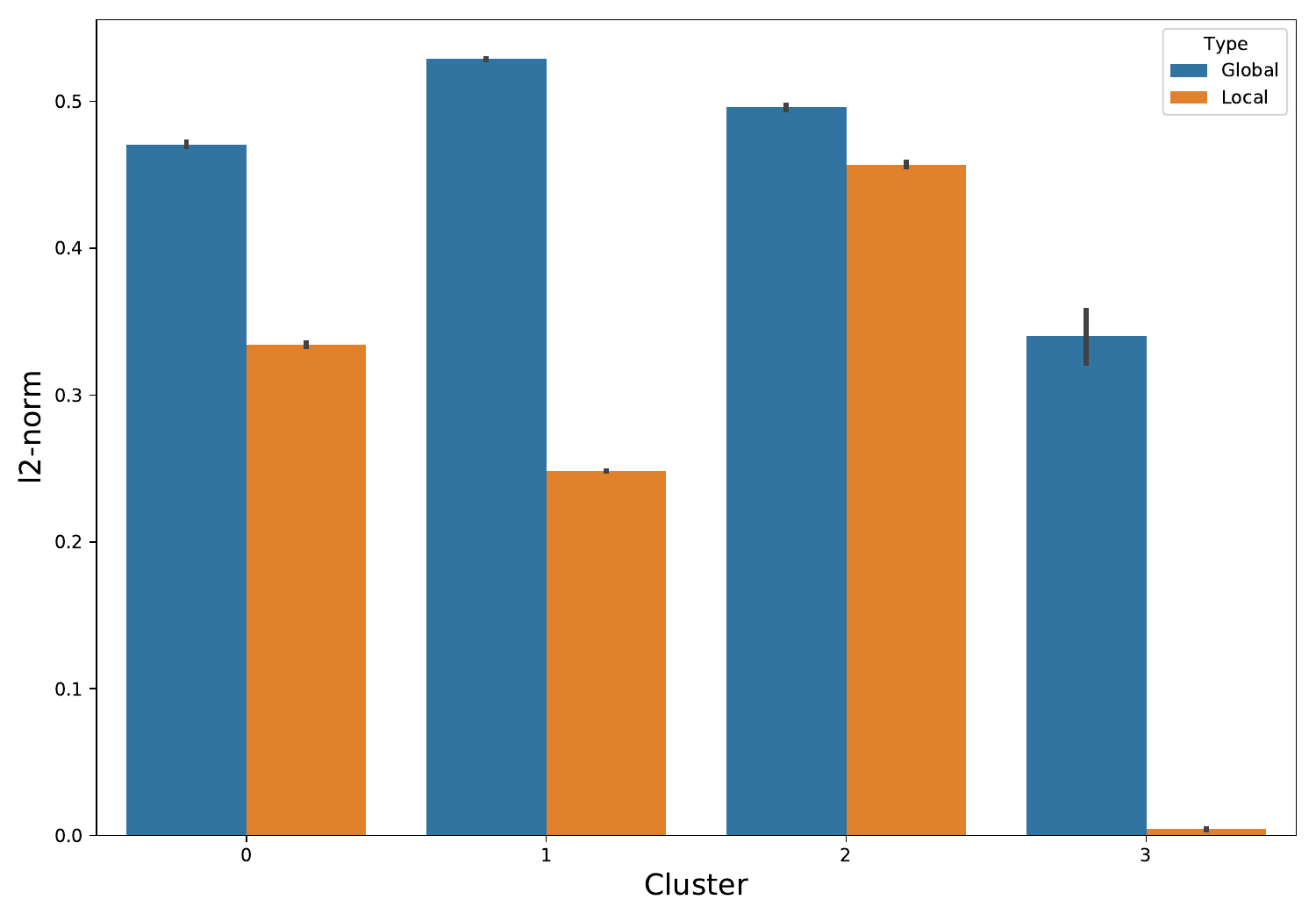}
\end{minipage}
\begin{minipage}[h]{0.49\linewidth}\centering\includegraphics[width=1\linewidth]{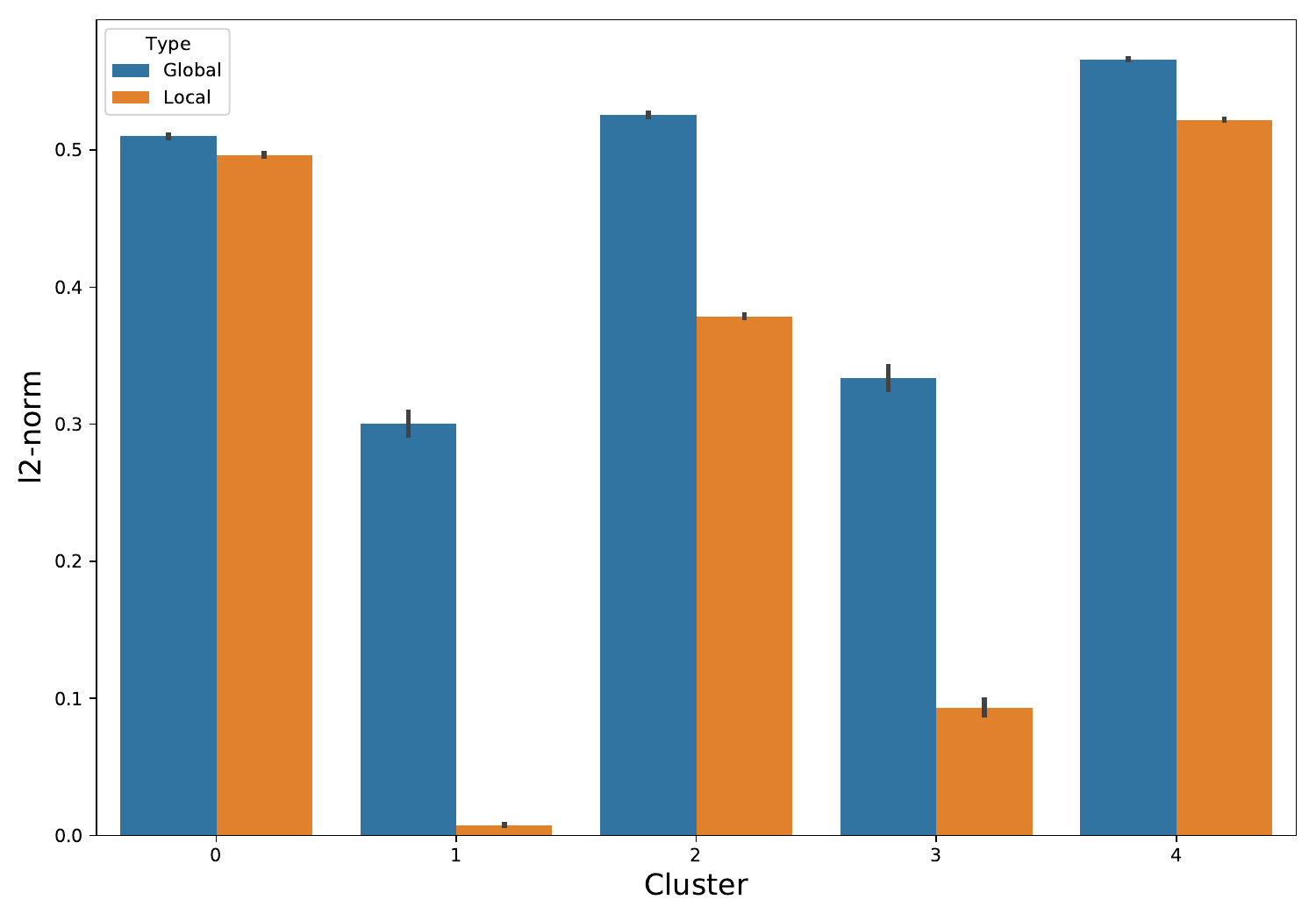}
\end{minipage}
\caption{Consistency analysis on LIME feature importance via inter-cluster distance measure, FBR with expert grouping (left) and FBR k-means clustering (right)}
\label{fig:LIME_real}
\end{center}
\end{figure}

Table~\ref{tab:fbr-reg-ess} showcases the performance comparison of the three approaches.
FBR with RF regressor on sub-populations grouped with k-means clustering algorithm, i.e. FBR(k)-RF, achieved the best performance out of the three types of approaches in forecasting heater energy consumption and outperformed (significantly) the traditional approach.
The performance of FBR with sub-population determined by domain experts, i.e. FBR(e), is on par with FBR(k), with slightly higher average errors (although the difference is not statistically significant).
Figure~\ref{fig:LIME_real} shows that the explanations provided using both local regression models, i.e. FBR, are significantly more consistent than those using a global model.

\begin{table*}[h]
\hspace{-1cm}
\caption{Regression performance for predicting heater energy consumption.}
\vspace{-0.35cm}
\begin{center}
\begin{tabular}{c|ccccc}
\toprule
& MAE & MSE & $R^2$ & MAPE \\ \midrule

RF &  0.4887  ±  0.0296 &  0.6012  ±  0.0777 &  0.3971  ±  0.0223 &  1.0243  ±  0.6831 \\
Ridge &  0.5889  ±  0.034 &  0.7057  ±  0.0648 &  0.2886  ±  0.0456 &  1.2453  ±  0.8305 \\
kNN &  0.5115  ±  0.0667 &  0.8359  ±  0.0974 &  0.1556  ±  0.098 &  1.4464  ±  0.7195 \\
MLP &  0.5715  ±  0.0653 &  0.795  ±  0.2184 &  0.2159  ±  0.125 &  1.758  ±  1.4801 \\

\midrule

FBR(e)-RF &  0.3538  ±  0.0227 &  0.2831  ±  0.0181 &  0.711  ±  0.0467 &  0.952  ±  0.6628 \\
FBR(e)-Ridge &  0.385  ±  0.0174 &  0.335  ±  0.0066 &  0.6582  ±  0.0485 &  0.9622  ±  0.4749 \\
FBR(e)-kNN &  0.3485  ±  0.0125 &  0.3449  ±  0.0119 &  0.649  ±  0.0432 &  1.1753  ±  0.5543 \\
FBR(e)-MLP &  0.4101  ±  0.0493 &  0.3859  ±  0.0693 &  0.6071  ±  0.083 &  1.1821  ±  0.4069 \\

\midrule

FBR(k)-RF &  0.3199  ±  0.0284 &  0.2585  ±  0.0367 &  0.7367  ±  0.0324 &  0.9141  ±  0.6433 \\
FBR(k)-Ridge &  0.4494  ±  0.0641 &  0.4686  ±  0.1377 &  0.5155  ±  0.172 &  1.2723  ±  0.9004 \\
FBR(k)-kNN &  0.3375  ±  0.04 &  0.3381  ±  0.042 &  0.6566  ±  0.0182 &  1.144  ±  0.5664 \\
FBR(k) MLP &  0.3793  ±  0.0343 &  0.348  ±  0.0602 &  0.6418  ±  0.0787 &  1.1354  ±  0.7279 \\

\bottomrule
\end{tabular}
\end{center}
\label{tab:fbr-reg-ess}
\end{table*}

\section{Conclusions and Future Work}

Accurate energy consumption prediction for electric commercial heavy-duty vehicles is an important AI task in which the models' explainability is paramount to gaining user trust.
In particular, a heterogeneous population that arises from varying configurations and usage patterns exhibits patterns similar to a well-known statistical phenomenon called Simpson’s paradox.
In this work, we illustrated that such a setting poses a challenge for XAI methods, yielding misleading results.
In this work, we demonstrated a divide-and-conquer-based approach (FBR) using a synthetic dataset consisting of heterogeneous sub-populations. Training multiple regression models on each sub-population instead of a global model on the entire dataset results in superior regression performance and more relevant and consistent LIME explanations. The same conclusion can be drawn from the experiment results on the real-world dataset, using FBR yields less error and more consistent explanations in predicting the energy consumption by the heaters.


Future works include: (i) developing an interactive mechanism to provide interpretable clustering results to the expert and receive supervision for improving the clustering setting (e.g., finding relevant contextual features for the given tasks); (ii) extending the experiments and the approach to work with time series data; (iii) conducting the study and the experiments on additional real-world datasets; (iv) explore whether different types of explanations exhibit varying resiliency to the issues presented in this work.

\section*{Acknowledgment}

The work was carried out with support from the Knowledge Foundation, Vinnova (Sweden's innovation agency) through the Vehicle Strategic Research and Innovation programme FFI, and the CHIST-ERA grant CHIST-ERA-19-XAI-012 funded by the Swedish Research Council.

%
%
\bibliographystyle{splncs04}
\bibliography{references}

\begin{thebibliography}{10}
\providecommand{\url}[1]{\texttt{#1}}
\providecommand{\urlprefix}{URL }
\providecommand{\doi}[1]{https://doi.org/#1}

\bibitem{InteractiveClusteringSurvey}
Bae, J., Helldin, T., Riveiro, M., Nowaczyk, S., Bouguelia, M.R., Falkman, G.: Interactive clustering: A comprehensive review. ACM Comput. Surv.  \textbf{53}(1) (feb 2020). \doi{10.1145/3340960}

\bibitem{bertsimas2018interpretable}
Bertsimas, D., Orfanoudaki, A., Wiberg, H.: Interpretable clustering via optimal trees (2018)

\bibitem{bertsimas2021interpretable}
Bertsimas, D., Orfanoudaki, A., Wiberg, H.: Interpretable clustering: An optimization approach. Mach. Learn.  \textbf{110}(1),  89–138 (jan 2021). \doi{10.1007/s10994-020-05896-2}, \url{https://doi.org/10.1007/s10994-020-05896-2}

\bibitem{dai2007Boosting}
Dai, W., Yang, Q., Xue, G.R., Yu, Y.: Boosting for transfer learning. In: Proceedings of the 24th International Conference on Machine Learning. p. 193–200. ICML '07, Association for Computing Machinery, New York, NY, USA (2007). \doi{10.1145/1273496.1273521}, \url{https://doi.org/10.1145/1273496.1273521}

\bibitem{fan2020transfer}
Fan, Y., Nowaczyk, S., R{\"o}gnvaldsson, T.: Transfer learning for remaining useful life prediction based on consensus self-organizing models. Reliability Engineering \& System Safety  \textbf{203},  107098 (2020)

\bibitem{gupta2021deep}
Gupta, V., Shi, H., Gimpel, K., Sachan, M.: Deep clustering of text representations for supervision-free probing of syntax (2021)

\bibitem{hardegree2006predicting}
Hardegree, S.P.: Predicting germination response to temperature. i. cardinal-temperature models and subpopulation-specific regression. Annals of Botany  \textbf{97}(6),  1115--1125 (2006)

\bibitem{HE2020118}
He, H., Khoshelham, K., Fraser, C.: A multiclass tradaboost transfer learning algorithm for the classification of mobile lidar data. ISPRS Journal of Photogrammetry and Remote Sensing  \textbf{166},  118--127 (2020). \doi{https://doi.org/10.1016/j.isprsjprs.2020.05.010}, \url{https://www.sciencedirect.com/science/article/pii/S0924271620301301}

\bibitem{li2016cluster}
Li, J., Weng, J., Shao, C., Guo, H.: Cluster-based logistic regression model for holiday travel mode choice. Procedia Engineering  \textbf{137},  729--737 (2016)

\bibitem{pan2010survey}
Pan, S.J., Yang, Q.: A survey on transfer learning. IEEE Transactions on knowledge and data engineering  \textbf{22}(10),  1345--1359 (2010)

\bibitem{scikit-learn}
Pedregosa, F., Varoquaux, G., Gramfort, A., Michel, V., Thirion, B., Grisel, O., Blondel, M., Prettenhofer, P., Weiss, R., Dubourg, V., Vanderplas, J., Passos, A., Cournapeau, D., Brucher, M., Perrot, M., Duchesnay, E.: Scikit-learn: Machine learning in {P}ython. Journal of Machine Learning Research  \textbf{12},  2825--2830 (2011)

\bibitem{verbeke2019fleet}
Verbeke, M., Murgia, A., Tourw{\'e}, T., Tsiporkova, E.: Fleet-based remaining useful life prediction of safety-critical electronic devices. In: Advances in Reliability, Risk and Safety Analysis with Big Data: Proceedings of the 57th ESReDA Seminar. pp. 23--24 (2019)

\bibitem{Wang2018Instance}
Wang, T., Huan, J., Zhu, M.: Instance-based deep transfer learning. CoRR  \textbf{abs/1809.02776} (2018), \url{http://arxiv.org/abs/1809.02776}

\bibitem{zhang2019quantile}
Zhang, Y., Wang, H.J., Zhu, Z.: Quantile-regression-based clustering for panel data. Journal of Econometrics  \textbf{213}(1),  54--67 (2019)

\end{thebibliography}

\end{document}